\documentclass{article}
\pdfoutput=1

\PassOptionsToPackage{numbers,compress}{natbib}

\usepackage[preprint]{neurips_2026}

\usepackage[utf8]{inputenc}
\usepackage[T1]{fontenc}
\usepackage{hyperref}
\usepackage{url}
\usepackage{booktabs}
\usepackage{amsmath}
\usepackage{amsfonts}
\usepackage{amssymb}
\usepackage{nicefrac}
\usepackage{microtype}
\usepackage{xcolor}
\usepackage{graphicx}
\usepackage{tikz}
\usetikzlibrary{arrows.meta, positioning, calc}

\newcommand{\E}{\mathbb{E}}
\newcommand{\R}{\mathbb{R}}

\newcommand{\Lsim}{\mathcal{L}_{\text{sim}}}
\newcommand{\Lpred}{\mathcal{L}_{\text{pred}}}
\newcommand{\Lvar}{\mathcal{L}_{\text{var}}}

\newcommand{\Lsigreg}{\mathcal{L}_{\text{SIGReg}}}
\newcommand{\Ltemporal}{\mathcal{L}_{\text{temporal}}}
\newcommand{\Lweak}{\mathcal{L}_{\text{Weak}}}

\newcommand{\Lvarhom}{\mathcal{L}_{\text{var-hom}}}
\newcommand{\Llateral}{\mathcal{L}_{\text{lateral}}}

\title{Local Synaptic Rules Can Implement a SIGReg Gradient Without Backpropagation}

\author{%
  Martin Andrews
}

\begin{document}

\maketitle

\begin{abstract}
We prove that two canonical local synaptic learning rules,
the potentiation arm of spike-timing-dependent plasticity (STDP$^+$)
and homeostatic plasticity (instantiated here via flashlight
granule-cell-like neurons), together can implement the exact gradient of a
SIGReg-like self-supervised learning objective.
The equivalence requires
  no gradient calculations, 
  no global error signals, 
  no weight transport, and 
  no label information: 
the only inputs are 
  pre- and post-synaptic firing rates, 
  local firing statistics, and the 
  temporal contiguity of natural sensory streams.
%
On a synthetic clustering task designed to probe whether
class structure can be recovered from temporal ordering of inputs alone,
ordered presentation raised cluster separation (CSR) to 2.49 while random
ordering left it near baseline (0.83), a roughly threefold ($\sim$3.5$\sigma$)
separation attributable solely to input ordering.
%
On temporally ordered MNIST, a two-layer network trained entirely with these rules 
achieved 87.3\% linear-probe accuracy, showing that the mechanism functions end-to-end.
%
%
\end{abstract}


\section{Introduction}

How does the brain learn useful representations without labels and without backpropagation?
The following two synaptic mechanisms are well replicated findings in systems neuroscience:
\begin{itemize}
\item{Spike-timing-dependent plasticity} (STDP) strengthens a synapse when the
pre-synaptic neuron fires shortly before the post-synaptic neuron, and weakens it
in the reverse order~\cite{bi_poo1998}.
\item{Homeostatic plasticity} regulates a neuron's long-run firing rate toward a
target distribution, preventing runaway excitation or silencing~\cite{turrigiano2008}.
\end{itemize}
Both rules are local, each depending only on quantities available at the synapse or the cell body.

Two key techniques for Self-supervised learning (SSL),
VICReg~\cite{bardes2022vicreg} and SIGReg~\cite{balestriero2025lejepaprovablescalableselfsupervised}, 
learn representations by maximising similarity between two views of the same input while
preventing dimensional collapse via variance and covariance regularisation.
While these objectives have been shown to be effective for unsupervised representation learning, 
they make no claims of neural plausibility.
The main result in this work addresses this, and in our model:
\begin{itemize}
\item The STDP weight gradient, applied to temporally-paired inputs, equals the
gradient of a $\Lpred$ similarity objective.
\item The flashlight homeostatic weight gradient equals the gradient of a weak-SIGReg
variance-covariance objective.
\end{itemize}

\begin{figure}[t]
  \centering
  \begin{tikzpicture}[
      >={Latex[length=2mm]},
      font=\small,
      neuron/.style={circle, draw, minimum size=6mm, inner sep=0pt},
      inp/.style={neuron, fill=black!6},
      enc/.style={neuron, fill=blue!12},
      encn/.style={neuron, fill=blue!12, dashed},
      fl/.style={neuron, fill=orange!22},
      inh/.style={rectangle, rounded corners, draw, fill=purple!12,
                  minimum height=6mm, inner sep=3pt, align=center},
      wsyn/.style={-, gray!55, line width=0.3pt},
      wshare/.style={-, gray!25, line width=0.2pt},
      fsyn/.style={->, black!55, line width=0.3pt},
      retro/.style={->, red!70!black, dashed, line width=0.8pt},
      lat/.style={->, purple!75!black, dash dot, line width=0.8pt},
      stdp/.style={->, blue!55!black, line width=0.9pt},
    ]
    \node[inp] (x1) at (0, 1.9) {\scriptsize$x_1$};
    \node[inp] (x2) at (0, 1.0) {\scriptsize$x_2$};
    \node[inp] (x3) at (0, 0.1) {\scriptsize$x_3$};
    \node[black!45, fill=white, inner sep=1pt] at (0,-0.75) {$\vdots$};
    \node[inp] (xn) at (0,-1.7) {\scriptsize$x_n$};
    \node[align=center, black!55] at (0,-2.7) {input\\stream $x$};
    \foreach \y/\l in {1.5/1, 0/2, -1.5/3}{\node[enc]  (ht\l) at (2.8,\y) {};}
    \foreach \y/\l in {1.5/1, 0/2, -1.5/3}{\node[encn] (hn\l) at (4.3,\y) {};}
    \node[blue!45!black] at (2.8,2.25) {$h_t$};
    \node[blue!45!black] at (4.3,2.25) {$h_{t+1}$};
    \node[align=center, blue!45!black] at (3.1,-2.7)
      {encoder $h{=}\sigma(Wx{+}b)$\\(same neurons at $t,\,t{+}1$)};
    \node[fl] (f1) at (7.4, 1.7) {\scriptsize$f_1$};
    \node[fl] (f2) at (7.4, 0.7) {\scriptsize$f_2$};
    \node[orange!55!black, fill=white, inner sep=1pt] at (7.4,-0.2) {$\vdots$};
    \node[fl] (fm) at (7.4,-1.5) {\scriptsize$f_m$};
    \node[align=center, orange!55!black] at (7.4,-2.85)
      {flashlight\\(granule)\\$f{=}hA^{\!\top}$};
    \foreach \a in {x1,x2,x3,xn}{\foreach \b in {ht1,ht2,ht3}{\draw[wsyn]   (\a) -- (\b);}}
    \foreach \a in {x1,x2,x3,xn}{\foreach \b in {hn1,hn2,hn3}{\draw[wshare] (\a) -- (\b);}}
    \node[gray!60!black, fill=white, inner sep=1pt] at (1.5,3.15) {$W$ (plastic)};
    \foreach \a in {hn1,hn2,hn3}{\foreach \b in {f1,f2,fm}{\draw[fsyn] (\a) -- (\b);}}
    \node[black!55, fill=white, inner sep=1pt] at (5.7,3.15) {$A$ (fixed)};
    \foreach \i in {1,2,3}{\draw[stdp] (ht\i) -- (hn\i);}
    \node[blue!55!black, fill=white, inner sep=1pt] at (3.6,1.9) {STDP$^+$};
    \node[text=red!70!black] at (8.3,2.6) {retrograde $(V_m{-}1)$};
    \draw[retro] (f1) to[out=118,in=35] (hn1);
    \draw[retro] (f2) to[out=132,in=12] (hn1);
    \node[inh] (bs) at (6.2,-4.0) {basket/stellate\\interneuron};
    \draw[fsyn, purple!70!black] (f1.east) to[out=-10,in=32] (bs.east);
    \draw[fsyn, purple!70!black] (f2.east) to[out=-15,in=16] (bs.east);
    \draw[fsyn, purple!70!black] (fm.east) to[out=-25,in=2]  (bs.east);
    \draw[lat] (bs.west) to[out=125,in=-62] (hn3);
    \node[text=purple!75!black] at (6.2,-4.7) {lateral $C_{mm'}$};
  \end{tikzpicture}
  \caption{Circuit realising the SIGReg gradient with local rules.
    The $h_t$ and $h_{t+1}$ strips are the \emph{same} encoder population at two
    consecutive timesteps (dashed = later step), not two hidden layers.
    Plastic encoder weights $W$ (grey, shared across both timesteps) are the
    only synapses that learn; the projection $A$ into the flashlight
    (granule-cell) population is fixed.
    Three local signals drive $W$:
    (i)~\textbf{STDP$^+$} (blue) compares the propensity vector across
    consecutive timesteps, $h_t\!\to\!h_{t+1}$, implementing the
    temporal-invariance term $\Lpred$;
    (ii)~a \textbf{retrograde} messenger (red, dashed) released by each
    flashlight neuron in proportion to its variance mismatch $(V_m-1)$,
    implementing the diagonal (variance) part of $\Lweak$;
    (iii)~\textbf{lateral inhibition} (purple, dash-dot) from basket/stellate
    interneurons driven by correlated flashlight pairs, implementing the
    off-diagonal (covariance) part $C_{mm'}$.
    No global error signal, weight transport, or labels are used in the circuit.
  }
  \label{fig:circuit}
\end{figure}
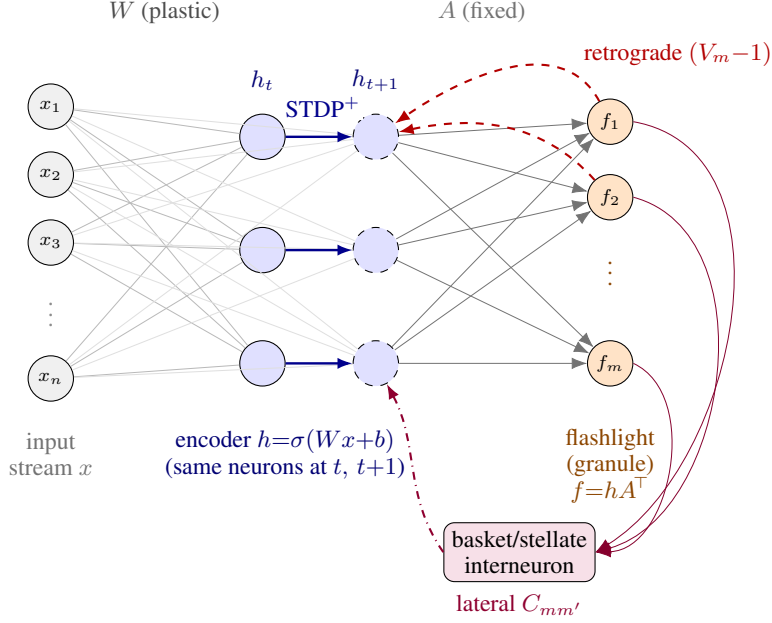

Together, STDP$^+$ and flashlight homeostasis implement gradient descent on the
full SIGReg objective without backpropagation or explicit data labels
(Figure~\ref{fig:circuit}).
We also investigated STDP$^-$, the depression (acausal) arm, which in principle
supplies the non-negativity projection that would make the combined system a
\emph{projected} gradient descent; on its own, however, it destabilises
learning, and a stable projection additionally requires an inhibitory
interneuron population that we leave to future work.

We emphasise that this equivalence is established by direct construction rather
than as a deep mathematical result: each biological rule, once written down, is
differentiated and shown to match a component of $\Lsigreg$ term by term. 
The contribution is the correspondence itself, namely that canonical,
independently-observed synaptic rules can compute exactly these gradients using only
locally available factors, not the elementary calculus that verifies it.
The mixed-sign random projection $A$ (see Section~\ref{section-flashlight}) 
is realised without violating Dale's Law, 
by splitting it into separate fixed excitatory (granule/mossy-fibre)
and inhibitory (Golgi-cell) populations that never change sign during learning.
Unlike prior local unsupervised methods that require an iterative inference
step (e.g.\ sparse coding's ISTA loop), the rules are strictly online and
single-pass: one forward activation, one local weight update, per input.
Code is provided\footnote{\url{https://github.com/mdda/biological-sigreg}} that
verifies all nine algebraic steps of this equivalence independently against JAX
autodiff~\cite{jax2018}.
%
%
%
%

\section{Background}

\subsection{Spike-Timing-Dependent Plasticity}

In this work, we model each neuron's activity as a \emph{propensity} --- 
its instantaneous firing probability under the Buesing~\cite{buesing2011} log-odds interpretation.
For a layer with weight matrix $W \in \R^{C \times n}$ and bias $b \in \R^C$,
the propensities $h \in \R^C$, one per neuron $i$, given input $x \in \R^n$ are
\begin{equation}
  h_i = \sigma\!\left(\textstyle\sum_j W_{ij}\,x_j + b_i\right), \qquad
  \sigma(u) = \tfrac{1}{1+e^{-u}}.
  \label{eq:propensity}
\end{equation}
The \emph{local gain} $\alpha_i = h_i(1-h_i) \in (0, \nicefrac{1}{4}]$ is
the slope of the neuron's f-I (firing-rate / input-current) curve at the current operating point,
and is maximal at $h_i = 0.5$ (maximally plastic), vanishing near saturation,
and is available at each neuron.

Introducing the notion of time into our model,
STDP operates asymmetrically on consecutive input pairs $(x_t, x_{t+1})$
with corresponding propensities $(h_t, h_{t+1})$, with distinct potentiation
and depression arms, denoted STDP$^+$ and STDP$^-$ respectively.

\textbf{STDP$^+$ (potentiation / LTP).}\;
When presynaptic neuron $j$ fires at time $t$ and postsynaptic neuron $i$
fires shortly after~\cite{bi_poo1998}, the synapse is strengthened:
\begin{equation}
  \Delta W_{ij}^{+}
  = +\E\!\left[\,\alpha_i(t)\;x_j(t)\;h_i(t{+}1)\,\right].
  \label{eq:stdp_plus}
\end{equation}
All three factors are locally available at the synapse:
$x_j(t)$ at the axon terminal,
$\alpha_i(t)$ at the soma via the slope of the f-I curve,
and $h_i(t{+}1)$ via a short-lived postsynaptic calcium trace or dendritic
delay line.

\textbf{STDP$^-$ (depression / LTD).}\;
When postsynaptic neuron $i$ fires at time $t$ before presynaptic neuron $j$
fires at $t{+}1$, the synapse is weakened:
\begin{equation}
  \Delta W_{ij}^{-}
  = -\E\!\left[\,h_i(t)\;\alpha_i(t{+}1)\;x_j(t{+}1)\,\right].
  \label{eq:stdp_minus}
\end{equation}
The three factors are again locally available: $h_i(t)$ via a decaying
postsynaptic calcium trace, $\alpha_i(t{+}1)$ at the soma, and $x_j(t{+}1)$
at the axon terminal.
Because $W_{ij} \geq 0$, depression is bounded: repeated LTD drives a synapse
to zero, effectively disconnecting it.
STDP$^-$ therefore acts as a \emph{causal filter} that prunes synapses
carrying acausal correlations (the postsynaptic neuron fired independently of
this presynaptic input), enforcing sparse, causally-selective connectivity.
It does \emph{not} implement a gradient step on the same objective as
STDP$^+$; we return to this distinction in Section~\ref{sec:equiv}.

\textbf{Forward prediction loss.}\;
The potentiation arm corresponds to gradient descent on a
\emph{forward prediction loss} that treats $h_{t+1}$ as a fixed teaching
signal --- a stop-gradient, written $\operatorname{sg}(\cdot)$:
\begin{equation}
  \Lpred(h_t,\,h_{t+1})
  = -\E\!\left[\,h_t \cdot \operatorname{sg}(h_{t+1})\,\right].
  \label{eq:lpred}
\end{equation}
Differentiating through $h_t$ only (since $\operatorname{sg}(h_{t+1})$ is held
constant):
\begin{equation}
  \frac{\partial \Lpred}{\partial W_{ij}}
  = -\E\!\left[\,\alpha_i(t)\;x_j(t)\;h_i(t{+}1)\,\right],
  \label{eq:grad_lpred}
\end{equation}
so $\Delta W_{ij}^{+} = -\partial \Lpred / \partial W_{ij}$ exactly:
\emph{STDP$^+$ is gradient descent on $\Lpred$}.

The stop-gradient is biologically natural: the synapse is updated at time $t$
using $h_{t+1}$ as an \emph{observed} result, not as a quantity whose
dependence on $W$ is tracked back through the network.
This mirrors BYOL~\cite{grill2020byol} in self-supervised learning,
where the target representation is held fixed when computing the gradient of
the online network.

\subsection{SIGReg and VICReg}

Turning now to the SSL literature, 
VICReg~\cite{bardes2022vicreg} prevents dimensional collapse by penalising the
full sample covariance matrix of the representation.
However, it is difficult to conceive of a biological mechanism for acting across such a covariance matrix.
The more recent SIGReg~\cite{balestriero2025lejepaprovablescalableselfsupervised} replaces
the full covariance with random projections.
Let $A \in \R^{M \times C}$ be a fixed matrix of $M$ random directions (rows
unit-normalised), drawn once and never updated, and write $f = hA^\top$ for
the projected representation. The collapse-prevention term (the
covariance-only \emph{weak} variant~\cite{akbar2026weaksigreg}) penalises the
deviation of the projected covariance from identity:
\begin{equation}
  \Lweak(f) = \bigl\|\operatorname{Cov}(f) - I\bigr\|_F^2.
  \label{eq:lweak}
\end{equation}
Then
\begin{equation}
  \Lsigreg
  = \underbrace{\Lpred(h_t, h_{t+1})}_{\text{temporal invariance}}
  + \lambda\,\underbrace{\Lweak(f)}_{\text{collapse prevention}}.
  \label{eq:sigreg}
\end{equation}
Here $h$ plays the role of the embedding $z$ in VICReg/SIGReg;
we retain $h$ to emphasise its biological interpretation as a propensity (firing-probability) vector.
The random-projection covariance $\operatorname{Cov}(f) = A\operatorname{Cov}(h)A^\top$
gives a low-rank view of the full covariance: since
$\operatorname{Var}(a_m\cdot h) = a_m^\top\operatorname{Cov}(h)\,a_m$,
constraining these projected variances across enough random directions pins down
$\operatorname{Cov}(h)$ itself, exactly so as $M \to \infty$.

\subsection{Flashlight homeostatic plasticity}
\label{section-flashlight}

Motivated by the effectiveness of these SSL methods,
we instantiate the random projection $A$ of Eq.~\eqref{eq:sigreg} as a second
population of $M$ \emph{flashlight} neurons, each computing one row of
$f = hA^\top$:
\begin{equation}
  f_m = a_m \cdot h, \qquad a_m \in \R^C,\ \|a_m\| = 1.
  \label{eq:flashlight}
\end{equation}
Biologically these could correspond to cerebellar granule cells, which receive
sparse random mossy-fibre input and are not subject to long-term synaptic
plasticity~\cite{marr1969,albus1971}.
The projection directions $a_m$ carry both positive and negative entries, 
consistent with granule cells receiving mixed 
excitatory (mossy-fibre) and fixed inhibitory (Golgi-cell) input.
This enables a net mixed-sign projection without violating
Dale's Law, since neither population changes its sign during learning.

Each flashlight neuron emits a homeostatic signal proportional to its
variance mismatch, driving plasticity at the upstream encoder weights $W$
(not $a_m$, which are fixed):
The \emph{afferent} synapses connecting the flashlight neurons
are protected from long-term plasticity (consistent with Marr/Albus),
while the homeostatic mechanism acts retrogradely,
reshaping the encoder representation until the flashlight probes report
$\operatorname{Cov}(h) \approx I$.

Within our model, we also include a second, 
inhibitory channel that suppresses cross-flashlight covariance via
local interneurons (perhaps corresponding to basket/stellate cells).
These two channels correspond to the diagonal and off-diagonal terms of
$\Lweak$ (Eq.~\ref{eq:lweak}):
\begin{align}
  \Lvarhom(f) &=
    \textstyle\sum_m \bigl(\operatorname{Var}(f_m) - 1\bigr)^2,
    \label{eq:lhom} \\
  \Llateral(f) &=
    \textstyle\sum_{m \neq m'} \operatorname{Cov}(f_m,f_{m'})^2.
    \label{eq:llateral}
\end{align}
Summing the two recovers $\Lweak$ exactly:
\begin{equation}
  \Lvarhom(f) + \Llateral(f) = \Lweak(f),
  \label{eq:lweak-decomp}
\end{equation}
since the Frobenius norm of $(\operatorname{Cov}(f) - I)$ splits into diagonal
and off-diagonal squared entries without remainder.
At the fixed point, $\operatorname{Cov}(f) = I$. As $M \to \infty$ the
projected-variance constraints $\operatorname{Var}(a_m\cdot h) = 1$ span all
1D directions, forcing $\operatorname{Cov}(h) = I$, i.e.\ $h$ is whitened
(second-order isotropic), which is the collapse-prevention condition of
weak-SIGReg~\cite{akbar2026weaksigreg}.
Full distributional Gaussianity is \emph{not} implied: the covariance-only
weak variant used here constrains second moments alone and does not test the
higher moments that a full Gaussianity criterion would.

\section{STDP$^+$ + Homeostasis $\equiv$ SIGReg Gradient; STDP$^-$ as Sparsifying Projection}
\label{sec:equiv}

Both STDP$^+$ and flashlight homeostasis share a common three-factor Hebbian
structure.
For a synapse $W_{ij}$ (from presynaptic neuron $j$ to postsynaptic neuron $i$),
each gradient-implementing update takes the form
\begin{equation}
  \Delta W_{ij} \;\propto\; \alpha_i \cdot \delta_i \cdot x_j,
  \label{eq:threefactor}
\end{equation}
where $x_j$ is the presynaptic activity (available at the axon terminal),
$\alpha_i = h_i(1-h_i)$ is the local gain (available at the soma), and
$\delta_i$ is a postsynaptic modulatory signal whose source differs between
rules.
We show that choosing $\delta_i$ appropriately in each case yields exactly
the gradient of a component of $\Lsigreg$.

STDP$^-$ plays a distinct role. 
Rather than implementing a gradient step, 
it acts on the non-negativity constraint $W_{ij} \geq 0$ 
that would, in principle, make the combined system a projected gradient descent on $\Lsigreg$.
On its own, however, it destabilises learning, as we show below.

\subsection{STDP\texorpdfstring{$^+$}{+} as the gradient of
  \texorpdfstring{$\Lpred$}{L\_pred}}

Equation~\eqref{eq:grad_lpred} already established that
\begin{equation}
  \frac{\partial \Lpred}{\partial W_{ij}}
  = -\E\!\left[\,\alpha_i(t)\;x_j(t)\;h_i(t{+}1)\,\right],
\end{equation}
so $\Delta W_{ij}^{+} = -\partial\Lpred/\partial W_{ij}$ exactly
(Eq.~\eqref{eq:stdp_plus}).
In the three-factor form~\eqref{eq:threefactor}, the postsynaptic signal is
$\delta_i = h_i(t{+}1)$: the propensity of neuron $i$ one step later,
accessible via a short-lived calcium trace or dendritic delay line.

\subsection{Flashlight homeostasis as the gradient of \texorpdfstring{$\Lweak$}{L\_Weak}}

We trace the gradient of $\Lweak$ back to the encoder weights
$W$ through three applications of the chain rule.

\textbf{Step 1: loss identity.}\;
From Eq.~\eqref{eq:lweak-decomp},
$\Lweak = \Lvarhom + \Llateral$
exactly.
Differentiation is therefore linear:
$\partial\Lweak/\partial f = \partial\Lvarhom/\partial f +
\partial\Llateral/\partial f$.

\textbf{Step 2: retrograde signal.}\;
Since $f = h\,A^\top$ and $A$ is fixed,
$\partial\Lweak/\partial h = (\partial\Lweak/\partial f)\,A$.
Expanding each component explicitly (using $\hat{f} = f - \E[f]$,
$V_m = \E[\hat{f}_m^2]$, $C_{mm'} = \E[\hat{f}_m\hat{f}_{m'}]$):
\begin{align}
  \frac{\partial \Lvarhom}{\partial h_i^{(n)}}
  &= \frac{4}{N} \sum_m (V_m - 1)\,\hat{f}_m^{(n)}\,a_{mi},
  \label{eq:retro_var} \\
  \frac{\partial \Llateral}{\partial h_i^{(n)}}
  &= \frac{4}{N} \sum_{m} \Bigl(\textstyle\sum_{m'\neq m} C_{mm'}\,\hat{f}_{m'}^{(n)}\Bigr)\,a_{mi}.
  \label{eq:retro_lat}
\end{align}

\textbf{Potential Biological pathways.}\;
Eq.~\eqref{eq:retro_var} is a \emph{retrograde signal} 
(e.g. endocannabinoid or nitric oxide) 
released by each flashlight neuron $m$ in proportion to its variance mismatch $(V_m - 1)$.
Equation~\eqref{eq:retro_lat} is the lateral inhibitory current arriving at
encoder neuron $i$ from basket/stellate cells driven by correlated granule-cell pairs.

\textbf{Step 3: weight gradient.}\;
Applying the chain rule through $h = \sigma(Wx + b)$:
\begin{equation}
  \frac{\partial \Lweak}{\partial W_{ij}}
  = \E\!\left[\,\alpha_i\cdot
    \frac{\partial\Lweak}{\partial h_i}\cdot x_j\,\right].
  \label{eq:hom_weight_grad}
\end{equation}

This is the three-factor Hebbian rule~\eqref{eq:threefactor} with
$\delta_i = \partial\Lweak/\partial h_i$ as the retrograde signal
(note that all three factors ($\alpha_i$, $\delta_i$, $x_j$) are locally available at or near synapse $W_{ij}$).
Unlike the fast, instantaneous STDP$^+$ factor, however, $\delta_i$ depends on
population statistics ($V_m$ and $C_{mm'}$) accumulated over many samples: it is
a slow, time-averaged signal, consistent with the comparatively slow timescale
of homeostatic plasticity.

\textbf{Combined result.}\;
The gradient-implementing updates sum to
\begin{equation}
  \Delta W^{+} + \lambda\,\Delta W^{\text{hom}}
  = \underbrace{-\frac{\partial\Lpred}{\partial W}}_{\text{STDP}^+}
  - \lambda\underbrace{\frac{\partial\Lweak}{\partial W}}_{\text{homeostasis}}
  = -\frac{\partial\,\Lsigreg}{\partial W}.
  \label{eq:total}
\end{equation}
Thus STDP$^+$ and flashlight homeostasis together implement gradient descent
on $\Lsigreg$ without any global error signal, weight transport, or label
information.
STDP$^-$ is not part of this gradient; its distinct role, and why it
destabilises learning on its own, is addressed next.

\subsection{STDP\texorpdfstring{$^-$}{-}: Sparsifying Projection}

STDP$^-$ does not correspond to any term in $\Lsigreg$.
Instead, it enforces the biological constraint $W_{ij} \geq 0$ through
continuous depression of synapses whose postsynaptic neuron fires
\emph{before} the relevant presynaptic input --- acausal correlations that
the network should not rely on.

Viewed through the lens of constrained optimisation, the combined action of
STDP$^+$ and STDP$^-$ on a non-negative weight would, in principle, be
equivalent to a \emph{projected gradient} update:
\begin{equation}
  W_{ij} \;\leftarrow\; \max\!\bigl(0,\;
    W_{ij} + \Delta W_{ij}^{+} + \Delta W_{ij}^{-}\bigr),
  \label{eq:proj_update}
\end{equation}
where the floor at zero is imposed by the physical impossibility of a
negative excitatory synapse rather than by an explicit algorithmic clip.
Because STDP$^-$ and STDP$^+$ operate simultaneously and continuously (not
as sequential steps), the projection is not a discrete post-hoc operation
but an emergent property of the competing plasticity arms.

The natural consequence is \emph{sparsification}: synapses that are never
causally recruited are depressed to zero and disconnected,
leaving each neuron connected only to inputs that it reliably predicts.
In our experiments, however, this pruning is too aggressive on its own.
The experiments below therefore exclude STDP$^-$ from the main conditions, 
and we return to its behaviour and the inhibitory population it would require in the Discussion.

\section{Experiments}
\label{sec:experiments}

\subsection{Synthetic clustering}

We construct a 3-class dataset of 450 samples in $D=50$ dimensions, where
the class structure lives in a 2D signal subspace embedded in 48 dimensions
of i.i.d.\ Gaussian noise (random-projection SNR $\approx 0.014$).
No labels are passed to the network at any point.
We evaluate the \emph{cluster separation ratio}
$\text{CSR} = d_{\text{between}} / d_{\text{within}}$,
where $\text{CSR} > 1$ indicates genuinely class-structured representations.

Under class-contiguous temporal presentation (the stream is ordered into
contiguous same-class blocks, so consecutive pairs are almost always same-class),
CSR rises from $0.73 \pm 0.25$ to $2.49 \pm 0.41$ over 300 epochs (mean $\pm$ std over 5 seeds).
Under random presentation order, the identical network never develops class structure: 
CSR stays near its initial value ($0.75 \pm 0.12$ to $0.83 \pm 0.23$).
The ordered network thus ends at CSR $= 2.49 \pm 0.41$ against $0.83 \pm 0.23$
for random order, a roughly threefold separation (about 3.5 standard deviations)
attributable solely to input ordering, since the two conditions differ in nothing else.
%

We observe that update magnitudes decay over training as homeostatic variance signals diminish.  
Within-class $\|\Delta W\|_{\text{within}}$ falls from $1.46 \pm 0.05$
to $0.52 \pm 0.04$ as $\operatorname{Cov}(f)\to I$. 
Boundary updates decay more slowly and remain comparable to (or somewhat above) within-class updates in later epochs 
(late-epoch boundary-to-within ratio $\approx 1.0$--$1.4$ across seeds),
reflecting the larger representation shift at class transitions.

\subsection{MNIST linear probe}

We train a two-layer network ([256, 128]) on MNIST using only STDP and
homeostatic plasticity, with no labels at any stage of representation learning.
Pairs are drawn from the class-contiguous temporal stream described above
(labels are used only to construct which pairs exist, never passed to the
network), and minibatches are randomly reordered before each weight update
so that the homeostatic term's per-batch covariance estimate reflects the
full class distribution rather than a single run.
After training, representations are frozen and a logistic regression head is
fitted on labelled examples (the standard SSL linear probe protocol).

\begin{table}[ht]
  \caption{MNIST linear probe and nearest-centroid accuracy.
    Architecture: [256, 128], 200 epochs. Values are mean $\pm$ std over
    5 seeds, except Raw pixels (deterministic).
    A$^{+-}$ reports the best of four STDP$^-$ learning rates; 1e-4 was best.
    A$^\text{b}$ and A$^{+-\text{b}}$ use half-normal (non-negative) weight initialisation.}
  \label{tab:mnist}
  \centering
  \small
  \begin{tabular}{llrr}
    \toprule
    Condition & Description & Probe    & Nearest-Centroid \\
              &             & Accuracy & Accuracy \\
    \midrule
    Raw pixels          & Logistic regression, 784-dim input         & 92.5\%          & ---    \\
    \midrule
    A                   & Full model: STDP$^+$ + homeostasis          & \textbf{87.3\% $\pm$ 0.4} & \textbf{77.7\% $\pm$ 1.0} \\
    B                   & Ablation: no homeostasis                    & 18.5\% $\pm$ 0.3          & 18.5\% $\pm$ 0.0 \\
    C                   & Ablation: random presentation order         & 86.5\% $\pm$ 0.6          & 64.9\% $\pm$ 1.8 \\
    \midrule
    A$^\text{b}$        & STDP$^+$ + non-negative weight init        & 87.1\% $\pm$ 0.3          & 77.1\% $\pm$ 1.0 \\
    A$^{+-}$            & STDP$^-$ + $W{\geq}0$ clipping             & 59.3\% $\pm$ 2.3          & 27.9\% $\pm$ 2.1 \\
    A$^{+-\text{b}}$    & STDP$^-$ + $W{\geq}0$ + non-neg init       & 69.6\% $\pm$ 2.5          & 38.7\% $\pm$ 3.4 \\
    \bottomrule
  \end{tabular}
\end{table}

The full model (A) reaches 87.3\% probe accuracy using only STDP$^+$,
homeostatic plasticity, and the temporal structure of the input stream ---
no labels, no backpropagation, no global error signal.
Condition B (no homeostasis) collapses to near-chance, 
indicating that homeostasis is important for preventing dimensional collapse.
Condition C (random order) establishes a whitening floor: 86.5\% probe
from homeostasis alone, without exploiting temporal contiguity.
Temporal ordering adds 0.8pp over this floor, consistent across all 5 seeds
(paired by seed, A exceeds C in every run).

Condition A$^\text{b}$ initialises all weights non-negative (half-normal,
variance-matched); since STDP$^+$ can only increase weights, they stay
non-negative throughout. Its 87.1\% probe accuracy is indistinguishable from
A, so the implicit inhibition from negative weights in Gaussian initialisation
is not required: homeostatic decorrelation suffices.
Conditions A$^{+-}$ and A$^{+-\text{b}}$ add STDP$^-$ with $W_{ij}\geq 0$
clipping, on Gaussian and non-negative initialisation respectively. Both drop
sharply (59.3\% and 69.6\%) with markedly higher seed variance
($\pm$2.3--2.5pp vs.\ $\leq$0.6pp elsewhere); we discuss why STDP$^-$ hurts,
and the inhibitory population that would remedy it, in the Discussion.

The raw-pixel baseline (92.5\%) exceeds all learned conditions.
This is expected: the network compresses 784 dimensions to 128 ($6\times$)
without any reconstruction objective,
and the compression cost exceeds the unsupervised structure recovered.


\section{Related Work}

\paragraph{Relation to prior biologically plausible methods.}
Several prior methods achieve high MNIST accuracy with local rules, 
but under fundamentally different evaluation protocols.
In terms of how the labels of examples are presented to the models: 
Forward-Forward~\cite{hinton2022ff} embeds the class label directly in the
input pixels; and 
Equilibrium Propagation~\cite{scellier2017} and
Predictive Coding~\cite{whittington2017pc} clamp output neurons to supervised targets during training.
Thus, these methods replace backpropagation as a \emph{credit assignment} mechanism
but are not unsupervised, since labels participate in every weight update.

The closest genuinely unsupervised analogue is single-layer feature learning
with an unsupervised dictionary followed by a linear classifier~\cite{coates2011sc},
which matches our evaluation protocol and reaches state-of-the-art accuracy on
natural-image benchmarks such as CIFAR-10 and NORB.
When the encoder is sparse coding, however, each input's code is computed by
solving an $\ell_1$-penalised optimisation (an iterative inner loop with no
plausible single-neuron or single-synapse implementation).
Our rules avoid any such per-input inner loop: each input drives a single
forward activation and a single local weight update. The homeostatic term does
rely on covariance statistics ($V_m$, $C_{mm'}$) pooled across a minibatch,
but this is a slow running average of population activity, not an iterative
inference procedure re-solved for every input.

A complementary question 
--- \emph{what synaptic configurations can a biologically detailed neuron express?} --- 
has recently been addressed with a digital-twin surrogate approach~\cite{aizenbud2026}: 
a single cortical pyramidal cell, optimised via backpropagation through a differentiable DNN twin, 
solves 10-bit parity and naturalistic sensory tasks well beyond what a perceptron achieves.
That work explicitly leaves open how such configurations might be learned by local rules; 
the equivalence proved here identifies STDP and homeostasis as candidate
mechanisms that could discover them without any global error signal.

\paragraph{Relation to the temporal derivative framework.}
A concurrent preprint~\cite{oreilly2026cortlearn} argues that neocortical learning
requires a \emph{temporal derivative}: the difference in neural activity between a
prediction phase and a subsequent outcome phase within each theta-cycle, implemented
via corticothalamic circuits and competing CaMKII/DAPK1 kinases.
The temporal intuition is shared with the present work --- both frameworks locate the
learning signal in activity \emph{changes} across time rather than activity magnitude,
and neither requires dedicated error-signal neurons.
The key distinction is supervision: the temporal derivative model clamps an external
outcome during the plus phase; the rules derived here need only the temporal contiguity
of the natural sensory stream.
The preprint~\cite{oreilly2026cortlearn} also dismisses spike-pair STDP as computationally insufficient; 
the critique does not apply to the rate-coded propensity formulation used here, 
which compares firing-rate vectors across consecutive timesteps and is closer to the BCM rule,
which has established neurochemical grounding via NMDA calcium dynamics~\cite{bi_poo1998}.
The two frameworks suggest a complementary rather than competing division:
temporal-derivative error-driven learning may operate at supervised output stages 
while STDP + homeostasis extracts unsupervised structure in earlier pathways.

\section{Discussion}

\paragraph{Depth scaling and the whitening cascade.}
A known limitation of decorrelation-based objectives applied at every layer is that
$\operatorname{Cov}(f) = I$ is a fixed point of the homeostatic gradient
(and hence, via Cramér-Wold, approximately $\operatorname{Cov}(h) = I$ as well):
once a layer has whitened its output, the homeostatic signal vanishes and the next layer sees a structureless input.
Our three-layer results confirm this empirically, and more starkly than a mere
plateau: adding a third layer costs 2.4--2.9pp of probe accuracy relative to
the matched two-layer condition (e.g.\ the full model falls from
$87.3\% \pm 0.4$ to $84.7\% \pm 0.7$, mean $\pm$ std over 5 seeds, consistent
across every seed tested). Built on an input layer 2 has already whitened,
the third layer is not merely redundant; it applies a further lossy
transformation with little structure left to extract.
This is consistent with the role of local receptive fields: 
in the V1$\to$V2$\to$V4 cortical hierarchy~\cite{felleman_vanessen1991}, each area whitens within its own spatial scale,
leaving cross-scale correlations intact for the next layer.
Convolutional variants of STDP and homeostasis, where the weight update is a
cross-correlation over spatial positions, would operate at different scales per
layer and break the cascade naturally~\cite{kavukcuoglu2010}.

\paragraph{Temporal contiguity as the positive-pair signal.}
SSL methods such as SimCLR~\cite{chen2020simclr} construct positive pairs via
explicit data augmentation, which requires design choices about what invariances to enforce.
In contrast, our framework requires no such choices:
consecutive observations of the same object in a natural sensory stream
constitute a positive pair automatically~\cite{schmidthuber1992histcomp}.
Collapse is prevented not by contrastive negative pairs but by the homeostatic
variance/covariance term, so no explicit negative-sample construction or memory
bank is required.
Since the positive-pair signal is an emergent property of the temporal
structure of the world (rather than arising from batching / augmentation)
these methods might be applied to domains where standard SSL techniques have no simple parallels.

\paragraph{STDP$^-$ and the missing inhibitory population.}
Empirically, STDP$^-$ with $W_{ij}\geq 0$ clipping degrades the learned
representation. 
Conditions A$^{+-}$ and A$^{+-\text{b}}$ in Table~\ref{tab:mnist} fall well below the full model, 
and a control with non-negative initialisation but no STDP$^-$ (A$^\text{b}$) does not, 
so the loss is attributable to STDP$^-$ itself rather than to the non-negativity constraint. 
This matches the argument of Section~\ref{sec:equiv}: STDP$^-$
supplies not a gradient step but the non-negativity projection
$W_{ij}\geq 0$, and on its own it continuously depresses weights, 
fighting the homeostatic variance signal. 
Realising the projection stably needs a compensating inhibitory population: 
adding basket/stellate-cell interneurons alongside STDP$^-$ would implement 
the full projected gradient descent on $\Lsigreg$ over the non-negative orthant. 
We leave this to future work.

\paragraph{Conclusion.}
STDP and homeostatic decorrelation, here instantiated as flashlight neurons
but general to any fixed random projection $A$, are not merely
\emph{analogous} to a self-supervised objective:
They \emph{are} its (projected) gradient, when carefully allowed to interact together in a biologically plausible setting.
More broadly, showing that a form of SIGReg is biologically plausible lets us
add it to a toolkit of \emph{allowable operations}: primitives that larger
biologically-constrained systems can be assembled from directly, using
familiar components rather than re-deriving each mechanism from scratch.
%

\begin{ack}
Support for this research was provided by the Google AI Developer Programs team, 
including access to the Gemini models and GPUs/TPUs on Google Cloud Platform.
\end{ack}


{\small
\bibliographystyle{abbrvnat}
\bibliography{refs}

\begin{thebibliography}{20}
\providecommand{\natexlab}[1]{#1}
\providecommand{\url}[1]{\texttt{#1}}
\expandafter\ifx\csname urlstyle\endcsname\relax
  \providecommand{\doi}[1]{doi: #1}\else
  \providecommand{\doi}{doi: \begingroup \urlstyle{rm}\Url}\fi

\bibitem[Aizenbud et~al.(2026)Aizenbud, Beniaguev, Pnueli, Segev, and
  London]{aizenbud2026}
I.~Aizenbud, D.~Beniaguev, N.~Pnueli, I.~Segev, and M.~London.
\newblock What can a neuron compute.
\newblock \emph{bioRxiv}, 2026.
\newblock \doi{10.64898/2026.06.08.730984}.
\newblock URL
  \url{https://www.biorxiv.org/content/early/2026/06/09/2026.06.08.730984}.

\bibitem[Akbar(2026)]{akbar2026weaksigreg}
H.~Akbar.
\newblock Weak-{SIGReg}: Covariance regularization for stable deep learning,
  2026.
\newblock URL \url{https://arxiv.org/abs/2603.05924}.

\bibitem[Albus(1971)]{albus1971}
J.~S. Albus.
\newblock A theory of cerebellar function.
\newblock \emph{Mathematical Biosciences}, 10\penalty0 (1):\penalty0 25--61,
  1971.
\newblock ISSN 0025-5564.
\newblock \doi{https://doi.org/10.1016/0025-5564(71)90051-4}.
\newblock URL
  \url{https://www.sciencedirect.com/science/article/pii/0025556471900514}.

\bibitem[Balestriero and
  LeCun(2025)]{balestriero2025lejepaprovablescalableselfsupervised}
R.~Balestriero and Y.~LeCun.
\newblock {LeJEPA}: Provable and scalable self-supervised learning without the
  heuristics, 2025.
\newblock URL \url{https://arxiv.org/abs/2511.08544}.

\bibitem[Bardes et~al.(2022)Bardes, Ponce, and LeCun]{bardes2022vicreg}
A.~Bardes, J.~Ponce, and Y.~LeCun.
\newblock {VICReg}: Variance-invariance-covariance regularization for
  self-supervised learning.
\newblock In \emph{International Conference on Learning Representations
  (ICLR)}, 2022.
\newblock URL \url{https://arxiv.org/abs/2105.04906}.

\bibitem[Bi and Poo(1998)]{bi_poo1998}
G.-q. Bi and M.-m. Poo.
\newblock Synaptic modifications in cultured hippocampal neurons: Dependence on
  spike timing, synaptic strength, and postsynaptic cell type.
\newblock \emph{Journal of Neuroscience}, 18\penalty0 (24):\penalty0
  10464--10472, 1998.
\newblock ISSN 0270-6474.
\newblock \doi{10.1523/JNEUROSCI.18-24-10464.1998}.
\newblock URL \url{https://www.jneurosci.org/content/18/24/10464}.

\bibitem[Bradbury et~al.(2018)Bradbury, Frostig, Hawkins, Johnson, Katariya,
  Leary, Maclaurin, Necula, Paszke, Vander{P}las, Wanderman-{M}ilne, and
  Zhang]{jax2018}
J.~Bradbury, R.~Frostig, P.~Hawkins, M.~J. Johnson, Y.~Katariya, C.~Leary,
  D.~Maclaurin, G.~Necula, A.~Paszke, J.~Vander{P}las, S.~Wanderman-{M}ilne,
  and Q.~Zhang.
\newblock {JAX}: composable transformations of {P}ython+{N}um{P}y programs,
  2018.
\newblock URL \url{http://github.com/jax-ml/jax}.

\bibitem[Buesing et~al.(2011)Buesing, Bill, Nessler, and Maass]{buesing2011}
L.~Buesing, J.~Bill, B.~Nessler, and W.~Maass.
\newblock Neural dynamics as sampling: A model for stochastic computation in
  recurrent networks of spiking neurons.
\newblock \emph{PLOS Computational Biology}, 7\penalty0 (11):\penalty0 1--22,
  11 2011.
\newblock \doi{10.1371/journal.pcbi.1002211}.
\newblock URL \url{https://doi.org/10.1371/journal.pcbi.1002211}.

\bibitem[Chen et~al.(2020)Chen, Kornblith, Norouzi, and Hinton]{chen2020simclr}
T.~Chen, S.~Kornblith, M.~Norouzi, and G.~Hinton.
\newblock A simple framework for contrastive learning of visual
  representations.
\newblock In \emph{International Conference on Machine Learning (ICML)}, 2020.
\newblock URL \url{https://arxiv.org/abs/2002.05709}.

\bibitem[Coates and Ng(2011)]{coates2011sc}
A.~Coates and A.~Y. Ng.
\newblock The importance of encoding versus training with sparse coding and
  vector quantization.
\newblock In \emph{Proceedings of the 28th International Conference on
  International Conference on Machine Learning}, ICML'11, page 921–928,
  Madison, WI, USA, 2011. Omnipress.
\newblock ISBN 9781450306195.

\bibitem[Felleman and Van~Essen(1991)]{felleman_vanessen1991}
D.~J. Felleman and D.~C. Van~Essen.
\newblock Distributed hierarchical processing in the primate cerebral cortex.
\newblock \emph{Cerebral Cortex}, 1\penalty0 (1):\penalty0 1--47, 1991.
\newblock \doi{10.1093/cercor/1.1.1}.

\bibitem[Grill et~al.(2020)Grill, Strub, Altch\'{e}, Tallec, Richemond,
  Buchatskaya, Doersch, Pires, Guo, Azar, Piot, Kavukcuoglu, Munos, and
  Valko]{grill2020byol}
J.-B. Grill, F.~Strub, F.~Altch\'{e}, C.~Tallec, P.~H. Richemond,
  E.~Buchatskaya, C.~Doersch, B.~A. Pires, Z.~D. Guo, M.~G. Azar, B.~Piot,
  K.~Kavukcuoglu, R.~Munos, and M.~Valko.
\newblock Bootstrap your own latent a new approach to self-supervised learning.
\newblock In \emph{Proceedings of the 34th International Conference on Neural
  Information Processing Systems}, NIPS '20, Red Hook, NY, USA, 2020. Curran
  Associates Inc.
\newblock ISBN 9781713829546.
\newblock URL \url{https://arxiv.org/abs/2006.07733}.

\bibitem[Hinton(2022)]{hinton2022ff}
G.~Hinton.
\newblock The forward-forward algorithm: some preliminary investigations, 2022.
\newblock URL \url{https://arxiv.org/abs/2212.13345}.

\bibitem[Kavukcuoglu et~al.(2010)Kavukcuoglu, Ranzato, and
  LeCun]{kavukcuoglu2010}
K.~Kavukcuoglu, M.~Ranzato, and Y.~LeCun.
\newblock Fast inference in sparse coding algorithms with applications to
  object recognition, 2010.
\newblock URL \url{https://arxiv.org/abs/1010.3467}.

\bibitem[Marr(1969)]{marr1969}
D.~Marr.
\newblock A theory of cerebellar cortex.
\newblock \emph{The Journal of Physiology}, 202\penalty0 (2):\penalty0
  437--470, 1969.
\newblock \doi{https://doi.org/10.1113/jphysiol.1969.sp008820}.
\newblock URL
  \url{https://physoc.onlinelibrary.wiley.com/doi/abs/10.1113/jphysiol.1969.sp008820}.

\bibitem[O'Reilly(2026)]{oreilly2026cortlearn}
R.~C. O'Reilly.
\newblock This is how the neocortex learns, 2026.
\newblock URL \url{https://arxiv.org/abs/2606.08720}.

\bibitem[Scellier and Bengio(2017)]{scellier2017}
B.~Scellier and Y.~Bengio.
\newblock Equilibrium propagation: bridging the gap between energy-based models
  and backpropagation.
\newblock \emph{Frontiers in Computational Neuroscience}, 11:\penalty0 24,
  2017.
\newblock URL
  \url{https://www.frontiersin.org/journals/computational-neuroscience/articles/10.3389/fncom.2017.00024}.

\bibitem[Schmidhuber(1992)]{schmidthuber1992histcomp}
J.~Schmidhuber.
\newblock Learning complex, extended sequences using the principle of history
  compression.
\newblock \emph{Neural Computation}, 4\penalty0 (2):\penalty0 234--242, 03
  1992.
\newblock ISSN 0899-7667.
\newblock \doi{10.1162/neco.1992.4.2.234}.
\newblock URL \url{https://doi.org/10.1162/neco.1992.4.2.234}.

\bibitem[Turrigiano(2008)]{turrigiano2008}
G.~G. Turrigiano.
\newblock The self-tuning neuron: synaptic scaling of excitatory synapses.
\newblock \emph{Cell}, 135\penalty0 (3):\penalty0 422--435, 2008.
\newblock URL \url{https://doi.org/10.1016/j.cell.2008.10.008}.

\bibitem[Whittington and Bogacz(2017)]{whittington2017pc}
J.~C.~R. Whittington and R.~Bogacz.
\newblock An approximation of the error backpropagation algorithm in a
  predictive coding network with local {H}ebbian synaptic plasticity.
\newblock \emph{Neural Computation}, 29\penalty0 (5):\penalty0 1229--1262, 05
  2017.
\newblock ISSN 0899-7667.
\newblock \doi{10.1162/NECO_a_00949}.
\newblock URL \url{https://doi.org/10.1162/NECO_a_00949}.

\end{thebibliography}
}

\newpage
\appendix
\section{Claim statements and verification details}
\label{app:claims}

Each algebraic step of the equivalence presented in Section~\ref{sec:equiv} 
is verified independently by comparing
the closed-form biological expression to a \texttt{jax.grad} evaluation of the
corresponding loss, on randomly initialised weights and activations
(batch 64, $C=16$, $M=64$, $n=32$, float32); 
see \texttt{verify\_gradients.py} in the codebase at \url{https://github.com/mdda/biological-sigreg}.

\paragraph{Auxiliary losses (claims 2--3).}
Claim~1 concerns $\Lpred = -\E[h_t\cdot\operatorname{sg}(h_{t+1})]$, the
stop-gradient forward-prediction loss used throughout the paper. Claims~2--3
instead concern the \emph{symmetric} similarity loss
$\Lsim = -\E[h_t\cdot h_{t+1}]$, which shares $\Lpred$'s forward value but
differs in its weight gradient (it differentiates through both views). $\Lsim$
splits as $\Lsim = \Lvar + \Ltemporal$ with
$\Lvar = -\E[\lVert h_t\rVert^2]$ and $\Ltemporal = \Lsim - \Lvar$:
$\Lvar$ is the variance-maintaining component that survives at temporal
equilibrium ($h_t = h_{t+1}$), while $\Ltemporal$ is the cross-correlation
component that vanishes there (the quiescence property). This decomposition is
included as supporting structure, showing that the STDP signal carries a
variance-maintaining part alongside a temporal-prediction part.

\paragraph{The nine verified identities.}
\begin{enumerate}\itemsep2pt
\item STDP$^+$ weight update $= -\,\partial\Lpred/\partial W$.
\item $\Lsim = \Lvar + \Ltemporal$ (loss identity).
\item $\partial\Lsim/\partial W = \partial\Lvar/\partial W + \partial\Ltemporal/\partial W$ (gradient identity).
\item $\Lvarhom + \Llateral = \Lweak$ (loss identity: diagonal and off-diagonal split of $\lVert\operatorname{Cov}(f)-I\rVert_F^2$).
\item $\partial\Lvarhom/\partial f + \partial\Llateral/\partial f = \partial\Lweak/\partial f$ (gradient identity).
\item The retrograde variance signal of Eq.~\eqref{eq:retro_var} equals $\partial\Lvarhom/\partial h$ (exact, no missing factor).
\item The variance three-factor Hebbian update $= -\,\partial\Lvarhom/\partial W$.
\item The lateral-inhibition Hebbian update $= -\,\partial\Llateral/\partial W$.
\item The full homeostatic weight update $= -\,\partial\Lweak/\partial W$.
\end{enumerate}

Table~\ref{tab:claims} reports the maximum absolute error of each identity.

\begin{table}[ht]
  \caption{Nine equivalence claims verified numerically by comparing
           analytical (biological) expressions to \texttt{jax.grad} outputs
           on randomly initialised weights and activations (float32).
           Max absolute error is over all entries of the relevant tensor.
           Errors at exactly $0$ arise where the identity is purely algebraic.}
  \label{tab:claims}
  \centering
  \small
  \begin{tabular}{clr}
    \toprule
    \# & Statement & Max abs.\ err \\
    \midrule
    1 & STDP$^+$ update $= -\partial\Lpred/\partial W$                          & $0$ \\
    2 & $\Lsim = \Lvar + \Ltemporal$ (loss)                 & $0$ \\
    3 & $\partial\Lsim/\partial W = \partial\Lvar/\partial W + \partial\Ltemporal/\partial W$ & $3.0\times10^{-8}$ \\
    \midrule
    4 & $\Lvarhom + \Llateral = \Lweak$ (loss) & $1.1\times10^{-5}$ \\
    5 & $\partial\Lvarhom/\partial f + \partial\Llateral/\partial f = \partial\Lweak/\partial f$ & $9.3\times10^{-9}$ \\
    6 & Retrograde signal (Eq.~\ref{eq:retro_var}) $= \partial\Lvarhom/\partial h$ & $2.2\times10^{-8}$ \\
    7 & Three-factor Hebbian (var) $= -\partial\Lvarhom/\partial W$ & $7.5\times10^{-8}$ \\
    8 & Three-factor Hebbian (lat) $= -\partial\Llateral/\partial W$ & $8.4\times10^{-9}$ \\
    9 & Full homeostatic gradient $= -\partial\Lweak/\partial W$ & $1.1\times10^{-7}$ \\
    \bottomrule
  \end{tabular}
\end{table}


\end{document}